\documentclass[sigconf]{acmart}
\AtBeginDocument{%
  }

\setcopyright{acmlicensed}
\copyrightyear{2018}
\acmYear{2018}

\acmDOI{XXXXXXX.XXXXXXX}

\acmISBN{978-1-4503-XXXX-X/2018/06}

\begin{document}

\title{Easy-Poly: An Easy Polyhedral Framework For 3D Multi-Object Tracking}



\author{Peng Zhang}
\affiliation{
  \institution{East China Normal University}
  \city{Shanghai}
  \country{China}
}
\email{52205901027@stu.ecnu.edu.cn}

\author{Xin Li}
\affiliation{
  \institution{Shanghai Artificial Intelligence Laboratory}
  \city{Shanghai}
  \country{China}
}
\email{lixin@pjlab.org.cn}

\author{Xin Lin \footnotemark[1]}
\affiliation{
  \institution{East China Normal University}
  \city{Shanghai}
  \country{China}
}
\email{xlin@cs.ecnu.edu.cn}

\author{Liang He}
\affiliation{
  \institution{East China Normal University}
  \city{Shanghai}
  \country{China}
}
\email{lhe@cs.ecnu.edu.cn}



\begin{abstract}

Recent 3D multi-object tracking (3D MOT) methods mainly follow tracking-by-detection pipelines, but often suffer from high false positives, missed detections, and identity switches, especially in crowded and small-object scenarios. To address these challenges, we propose Easy-Poly, a filter-based 3D MOT framework with four key innovations: (1) CNMSMM, a novel Camera-LiDAR fusion detection method combining multi-modal augmentation and an efficient NMS with a new loss function to improve small target detection; (2) Dynamic Track-Oriented (DTO) data association that robustly handles uncertainties and occlusions via class-aware optimal assignment and parallel processing strategies; (3) Dynamic Motion Modeling (DMM) using a confidence-weighted Kalman filter with adaptive noise covariance to enhance tracking accuracy; and (4) an extended life-cycle management system reducing identity switches and false terminations. Experimental results show that Easy-Poly outperforms state-of-the-art methods such as Poly-MOT and Fast-Poly, achieving notable gains in mAP (e.g., from 63.30\% to 65.65\% with LargeKernel3D) and AMOTA (e.g., from 73.1\% to 75.6\%), while also running in real-time. Our framework advances robustness and adaptability in complex driving environments, paving the way for safer autonomous driving perception.


\end{abstract}




\maketitle
\renewcommand{\thefootnote}{\fnsymbol{footnote}} 
\section{Introduction}
 
\label{sec: introduction}

Recent advances in 3D multi-object tracking (3D MOT), driven by autonomous driving, robotics, and mixed reality, emphasize multi-modal data fusion to enhance reliability and accuracy~\cite{pftrack,ding20233dmotformer}. Traditional filtering and deep learning coexist, yet most methods prioritize tracking over improving 3D detection. For example, Poly-MOT and Fast-Poly~\cite{li2024fast} utilize FocalsConv~\cite{chen2022focal} with CenterPoint~\cite{yin2021center} and LargeKernel3D~\cite{chen2022scaling}, but lack thorough multi-modal data augmentation, NMS, and loss optimization. They suffer from high identity switches (IDS), false negatives (FN), and false positives (FP), especially in crowded and small-object scenarios, highlighting the need for integrated detection and tracking enhancements.

We propose \textbf{Easy-Poly}, a real-time filter-based 3D MOT under the \emph{tracking-by-detection} paradigm, supporting multiple categories with improved performance in complex settings. First, we develop a Combined NMS Framework for 3D Multi-Modal Small Object Detection (\textbf{CNMSMM}) based on FocalsConv, which significantly improves detection via refined multi-modal augmentation, efficient NMS, and a novel loss function, yielding superior results in dense traffic and small-object detection. Next, the optimized \emph{Easy-Poly} pipeline (Figure~\ref{fig: frameworkpoly}) manages detection flow through pre-processing, association, matching, estimation, and life-cycle control. Key components include \textbf{Score Filtering (SF)} and NMS to remove low-confidence detections, a \textbf{Dynamic Track-Oriented (DTO)} module for ambiguous association, and a \textbf{Dynamic Motion Modeling (DMM)} strategy that incorporates confidence-weighted Kalman filter updates with adaptive noise covariance tuning. Collectively, these innovations reduce false positives, tracking failures, and enhance robustness.

Extensive experiments on the nuScenes dataset validate the efficacy of Easy-Poly, demonstrating superior detection and tracking performance compared to Poly-MOT and Fast-Poly baselines. With the CNMSMM framework, \emph{CenterPoint} achieves an mAP improvement from \textbf{63.86\%} to \textbf{65.45\%}, while \emph{LargeKernel3D} improves from \textbf{63.30\%} to \textbf{65.65\%} (See the Table~\ref{tab:mAP}). Moreover, in 3D MOT, Easy-Poly increases the AMOTA of CenterPoint from \textbf{73.1\%} to \textbf{74.5\%}, and LargeKernel3D reaches \textbf{75.6\%} AMOTA at \textbf{34.9} FPS—almost doubling the inference speed of the baseline. Our framework thereby offers an efficient, versatile, and more reliable solution for real-world autonomous driving, underscoring the importance of synergistic improvements in both detection and tracking modules.

This paper makes the following key contributions:
\begin{itemize}
    \item We propose a novel CNMSMM 3D detection framework that integrates refined NMS/Loss methods and multi-modal data augmentation, significantly improving 3D detection quality in dense-traffic and small-object scenarios.
    \item We introduce an \textbf{Easy-Poly} pipeline encompassing pre-processing, data association, and motion modeling. Notably, we design a class-aware DTO algorithm for robust association under ambiguous conditions and a DMM module that adaptively adjusts noise covariances.
    \item We incorporate confidence-weighted Kalman filtering and dynamic threshold management into life-cycle management, enhancing tracking accuracy and robustness in small object and complex road conditions.
    \item Extensive experiments on nuScenes show that Easy-Poly outperforms Poly-MOT and Fast-Poly baselines in both detection (mAP, NDS) and MOT metrics (AMOTA, MOTA, FPS), validating the effectiveness and efficiency of our design.
\end{itemize}

\begin{table}
\caption{Comparison of existing 3D object detection methods applied to the nuScenes validation set. Except for the method in line 2, which utilizes only LiDAR data, all other methods employ multi-modal inputs.}
\label{tab:mAP}
    \setlength{\tabcolsep}{1.0mm}
  \begin{tabular}{ccclccl}
    \toprule
    \textbf{Framework} & \textbf{Method} & \textbf{Fusion} & \textbf{mAP} & \textbf{NDS} \\
    \midrule
     FocalsConv~\cite{chen2022focal} & CenterPoint~\cite{yin2021center} &  \checkmark & 63.86 & 69.41 \\

    FocalsConv~\cite{chen2022focal} & LargeKernel3D~\cite{chen2022scaling} & $\times$  & 60.30 & 67.50 \\

      FocalsConv~\cite{chen2022focal} & LargeKernel3D~\cite{chen2022scaling} & \checkmark & 63.30 & 69.10 \\
    \midrule
     \textbf{Ours} & CenterPoint~\cite{yin2021center} &  \checkmark & 
        \textcolor{blue}{65.45} & 
        \textcolor{blue}{70.29} \\
     \textbf{Ours} & LargeKernel3D~\cite{chen2022scaling} & \checkmark  & 
        \textcolor{red}{65.65} & 
        \textcolor{red}{70.48} \\
    \bottomrule
  \end{tabular}
\end{table}

\section{Related Work}
\label{sec: related work}

\subsection{3D Object Detection}
Recent advances in 3D object detection for autonomous driving are classified by sensor modality into \textbf{camera-based}, \textbf{point cloud-based}, and \textbf{multi-modality-based} methods \cite{3dodsurvey2022}. Camera-based approaches include monocular, multi-camera, BEV, and pseudo-LIDAR techniques. Point cloud methods cover voxel, point-voxel hybrids, BEV, 4D radar, spatiotemporal modeling, and data augmentation. Multi-modal methods emphasize deep fusion and cross-modal interactions to improve detection accuracy and robustness.

\noindent
\textbf{LIDAR-only Detectors} handle point clouds through (1) learning objectives such as anchor-based \cite{UTsemi2022} and anchor-free \cite{Afdet2020} methods; (2) data representations including point-based, grid-based, point-voxel hybrid, range-based, and 4D radar; and (3) auxiliary techniques like data augmentation \cite{Qdataaug2020}, spatiotemporal modeling, and pseudo-labeling. Although LIDAR offers superior speed and accuracy, point cloud sparsity causes missed detections, particularly for small or distant objects. Fusion with camera data effectively addresses these issues, enhancing detection completeness and reliability.

\noindent
\textbf{Multi-Modal Fusion} includes camera-radar, camera-LIDAR \cite{li2024fast,li2023poly}, camera-4D radar, LIDAR-4D radar, and LIDAR-radar combinations. Camera-LIDAR fusion is categorized into (1) early fusion merging raw data before feature extraction \cite{lang2019pointpillars,PointNet2018}; (2) deep fusion integrating multi-level features \cite{li2024fast,li2023poly}; and (3) late fusion combining high-level features or detections \cite{CLOCs2020,Fast-CLOCs2022}. Utilizing diverse modalities such as image, point cloud, millimeter-wave, depth, and 4D radar is essential for detection accuracy. We propose CNMSMM, a novel Camera-LIDAR fusion framework that employs advanced data augmentation and optimized Loss/NMS, achieving state-of-the-art 3D detection accuracy and robustness, especially in crowded and small-object scenarios.

\subsection{3D Multiple Object Tracking}

3D MOT is essential for autonomous driving perception, with notable progress since AB3DMOT~\cite{weng20203d} introduced a 3D extension of the \textbf{Tracking-by-Detection (TBD)} framework. Despite advances in occlusion handling, data association, and long-term tracking, methods such as Poly-MOT and Fast-Poly continue to face challenges in crowded scenes, small object tracking, and trajectory errors. Easy-Poly addresses these issues through innovations across the TBD pipeline, including pre-processing, estimation, association, and life-cycle management.

In pre-processing, Easy-Poly improves detection by integrating multi-modal data augmentation for LargeKernel3D and optimizing Non-Maximum Suppression (NMS) thresholds, significantly enhancing performance on small objects and crowded scenes, which prior works~\cite{pang2022simpletrack, li2023poly, li2023camo} have largely neglected. The estimation stage employs filters (Linear Kalman~\cite{kim2021eagermot, li2023camo, pang2022simpletrack, PC3T}, Extended Kalman~\cite{li2023poly}, Point Filter~\cite{benbarka2021score, yin2021center}) and motion models to predict and update tracklets. Easy-Poly advances AMOTA by optimizing BICYCLE parameters via grid search and Bayesian optimization, and incorporates detector confidence as a weighting factor in Kalman filtering (DMM), thereby enhancing tracking accuracy and robustness.

Previous Association work Association depends on geometry-based affinities (IoU~\cite{weng20203d, wang2022deepfusionmot}, GIoU~\cite{li2023camo, li2023poly, pang2022simpletrack}, Euclidean~\cite{benbarka2021score, PC3T, kim2021eagermot}, NN distance~\cite{ding20233dmotformer, sadjadpour2023shasta, zaech2022learnable, gwak2022minkowski}) and appearance cues. Although progress has been made, occlusions and false positives remain challenging. Easy-Poly’s DTO algorithm effectively resolves these ambiguities. Life-cycle management controls tracklet initialization, termination, and merging through count- or confidence-based strategies~\cite{kim2021eagermot, li2023camo, li2023poly, pang2022simpletrack, wang2022deepfusionmot, weng20203d, benbarka2021score, PC3T}. Easy-Poly extends tracklet lifespan and dynamically adjusts thresholds, reducing premature termination and improving robustness for sustained tracking.

\begin{figure*}[t]
    \centering
    \includegraphics[width=0.98\linewidth]{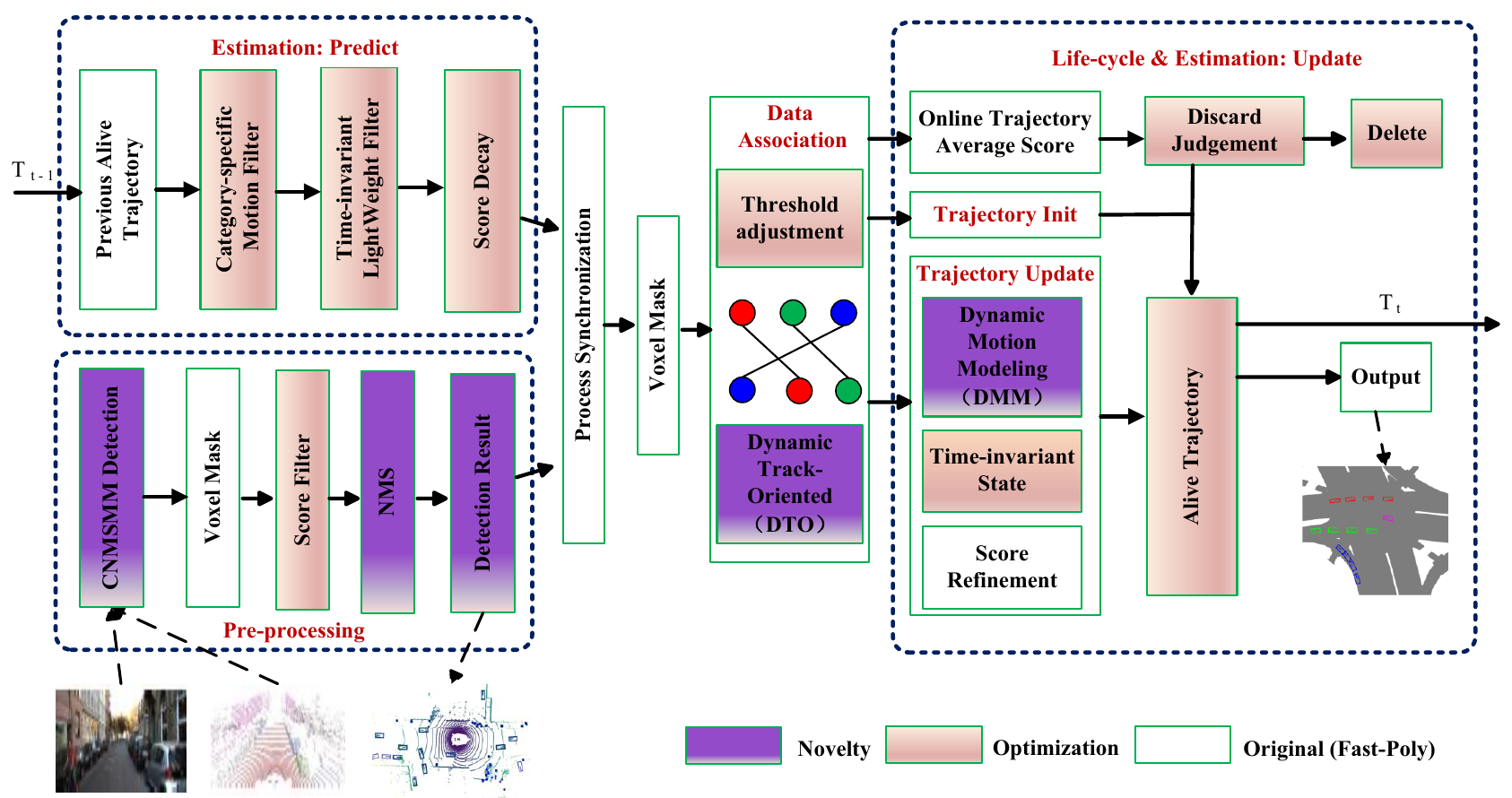}
    \vspace{-8pt}
    \caption
    {
        The pipeline of the Easy-Poly method is illustrated, with real-time enhancements over the baseline Fast-Poly highlighted using distinct colors. \textbf{\textcolor[RGB]{242, 156, 177}{Pink}} indicates optimizations applied to the Fast-Poly method, while \textbf{\textcolor[RGB]{138, 43, 226}{Purple}} represents newly introduced functional modules.
     }
     \Description{}
    \label{fig: frameworkpoly}
\end{figure*}

\section{Methodology}
\label{sec: method}

This work proposes \textbf{Easy-Poly}, a real-time 3D MOT method leveraging efficient filter-based algorithms within the TBD paradigm. The approach robustly manages multiple object categories and achieves substantial performance gains in complex environments, providing a practical and scalable solution for autonomous driving.

\subsection{Easy-Poly Framework}

The Easy-Poly framework substantially improves upon the previous Fast-Poly method by introducing multiple optimizations and innovations, as depicted in Figure~\ref{fig: frameworkpoly}.

The pipeline consists of five main stages: \textbf{Pre-processing}, \textbf{Association}, \textbf{Matching}, \textbf{Estimation}, and \textbf{Life-cycle}. The procedure proceeds as follows:

\textbf{Pre-processing and Prediction:} Utilizes outputs from the CNMSMM detection framework, formatting the initial frame for tracking. For each frame, Easy-Poly applies parallel processing to filter 3D detections and predict existing trajectories. SF and NMS filters refine detections, while specialized filters estimate motion (time-variable), score, and time-invariant trajectory states.

\textbf{Association and Matching:} Construct cost matrices for two-stage association. Voxel mask and geometry-based metrics accelerate NMS and matching computations. The DTO algorithm~\cite{kuhn1955hungarian} determines matched pairs, unmatched detections, and unmatched tracklets.

\textbf{State Update:} Optimizes matrix dimensionality by updating time-invariant and time-variable states in matched tracklets using an optimized lightweight filter and Extended Kalman Filter (EKF), respectively. Scores are refined via a confidence-count mixed life-cycle approach. A novel DMM module is introduced to enhance robustness under varying environmental conditions.

\textbf{Initialization and Termination:} Initialize unmatched detections as new active tracklets. False positive agents are identified through soft termination based on max-age and online average refined scores.

\textbf{Frame Advancement:} Forwards remaining tracklets to downstream tasks and prepares them for subsequent frame tracking.

\begin{figure*}[t]
    \centering
    \includegraphics[width=0.98\linewidth]{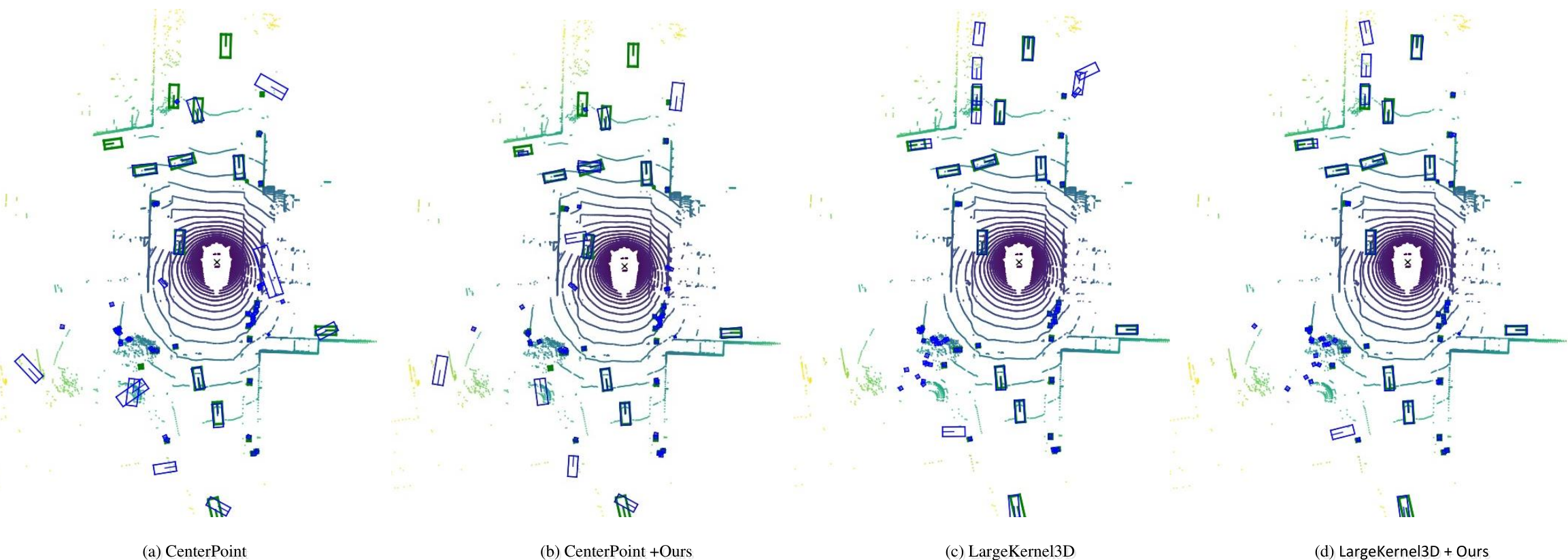}
    \vspace{-8pt}
    \caption{
     The detection performance of the CenterPoint and LargeKernel3D methods in multi-modal mode is compared between FocalsConv and CNMSMM. This comparison illustrates the performance improvements obtained by CNMSMM in 3D object detection, particularly for small objects. Figures \textbf{(a)} and \textbf{(c)} present results using the baseline FocalsConv model, whereas Figures \textbf{(b)} and \textbf{(d)} show results obtained with the CNMSMM framework. The results emphasize the superiority of CNMSMM over the original methods, especially in detecting small objects and handling crowded scenes.}
     \Description{}
    \label{fig: topfig}
\end{figure*}

\subsection{CNMSMM Framework}
 
We propose CNMSMM, a novel camera-LiDAR fusion detection framework that jointly optimizes efficient NMS and loss functions to enhance detection performance, particularly for small objects in complex scenes. Fig~\ref{fig: topfig} illustrates the superior 3D detection performance of CNMSMM compared to FocalsConv for CenterPoint and LargeKernel3D, with notable improvements in small object detection. CNMSMM reconstructs FocalsConv by integrating advanced augmentation techniques and a hybrid NMS/loss strategy to achieve effective fusion of image and point cloud data, while robustly handling erroneous and empty frames. As a scalable platform, it supports multiple modalities and models, including PointPillars, CenterPoint, and LargeKernel, enabling seamless integration. Key innovations include a dynamic adaptive NMS that combines Circle NMS and Gaussian Soft-NMS, a composite loss function (CIOU combined with Quality Focal Loss) tailored for small objects, and a robust, scalable network design. This framework improves indexing efficiency and detection accuracy during training, evaluation, and subsequent stages, significantly enhancing metrics such as mAP and NDS (Table~\ref{tab:mAP}). On the nuScenes dataset, the integration of LargeKernel3D within CNMSMM increases mAP by 2.35\% and NDS by 1.38\%, with substantial gains in small object detection.
Implemented using Det3D, CNMSMM incorporates a multimodal LargeKernel3D module and a novel VoxelSpecial module for efficient voxelization. The preprocessing stage employs a 3D ResNet backbone for voxelized LiDAR data and a ResNet50 backbone for RGB images. Multi-modal augmentation methods—including \textbf{db sampler}, \textbf{flip}, \textbf{rotate}, \textbf{rescale}, and \textbf{translate}—were analyzed, revealing that the combination of \textbf{double flip} and \textbf{rotation} significantly improves accuracy in crowded and small-object scenarios. The Point2ImageProjection module projects LiDAR voxels onto image coordinates, while the VoxelWithPointProjection module samples and fuses image features with LiDAR voxels. A region proposal network (RPN) generates 3D bounding boxes that are refined by the ROI Head. A hierarchical NMS, combining Circle NMS and Gaussian Soft-NMS, suppresses redundancies, followed by score threshold filtering. The loss optimization within this pipeline substantially enhances small object detection and overall system robustness.

\subsection{Dynamic Track-Oriented Data Association}

In the data association stage, we present the \textbf{DTO} algorithm, a novel method that complements traditional \textbf{Hungarian}, \textbf{Greedy}, and \textbf{Mutual Nearest Neighbor (MNN)} approaches. DTO, an enhanced variant of Multiple Hypothesis Tracking (MHT)~\cite{kim2015multiple}, mitigates uncertainties in object-tracking assignment by maintaining and updating multiple association hypotheses between detections and tracks over time. This method facilitates robust management of complex scenarios as new observations are incorporated. The Dynamic Track-Oriented (DTO) data association algorithm is formally defined as follows:

\begin{equation}
    \mathbf{m}_{\mathrm{det}}, \mathbf{m}_{\mathrm{tra}}, \mathbf{u}_{\mathrm{det}}, \mathbf{u}_{\mathrm{tra}} = \mathrm{DTO}(\mathbf{C}, \boldsymbol{\tau})
    \label{eq:DTO}
\end{equation}

$\mathbf{C} \in \mathbb{R}^{N_{\mathrm{cls}} \times N_{\mathrm{det}} \times N_{\mathrm{tra}}}$ is the cost matrix between classes, detections, and tracks. If the cost matrix has a size of $N_{\mathrm{det}} \times N_{\mathrm{tra}}$, then $\mathbf{C} \in \mathbb{R}^{N_{\mathrm{det}} \times N_{\mathrm{tra}}}$. $\boldsymbol{\tau} \in \mathbb{R}^{N_{\mathrm{cls}}}$ is the matching threshold for each class.  $\mathbf{m}_{\mathrm{det}} \in \mathbb{N}^{|\mathbf{m}_{\mathrm{det}}|}$ is the list of matched detection indices. $\mathbf{m}_{\mathrm{tra}} \in \mathbb{N}^{|\mathbf{m}_{\mathrm{tra}}|}$ is the list of matched track indices.  $\mathbf{u}_{\mathrm{det}} \in \mathbb{N}^{|\mathbf{u}_{\mathrm{det}}|}$ is the list of unmatched detection indices. $\mathbf{u}_{\mathrm{tra}} \in \mathbb{N}^{|\mathbf{u}_{\mathrm{tra}}|}$ is the list of unmatched track indices. 

Equation \eqref{eq:DTO} describes the main function of the DTO algorithm, which is to take the cost matrix $\mathbf{C}$ and the matching thresholds $\boldsymbol{\tau}$ as inputs, and output the lists of matched detection and track indices $\mathbf{m}_{\mathrm{det}}$ and $\mathbf{m}_{\mathrm{tra}}$, as well as the lists of unmatched detection and track indices $\mathbf{u}_{\mathrm{det}}$ and $\mathbf{u}_{\mathrm{tra}}$.

The algorithm proceeds as follows:

\begin{enumerate}

    \item For each class \( c \in \{1, \ldots, N_{\mathrm{cls}}\} \), extract the corresponding cost matrix slice:

    \begin{equation}
    \mathbf{C}_c = \mathbf{C}[c, :, :] \in \mathbb{R}^{N_{\mathrm{det}} \times N_{\mathrm{tra}}}
    \label{eq:DTO1}
    \end{equation}

    \item Solve the optimal assignment using the Hungarian algorithm:
 
    \begin{equation}
     \mathbf{M}_c = \mathrm{Hungarian}(\mathbf{C}_c)
    \label{eq:DTO2}
    \end{equation}
 
    where \(\mathbf{M}_c = \{(i,j) \mid i \in [0, N_{\mathrm{det}}-1], j \in [0, N_{\mathrm{tra}}-1]\}\) denotes matched detection-track pairs.
    
    \item Filter matches by the class-specific threshold \(\tau_c\):

    \begin{equation}
     \mathbf{M}_c^{\mathrm{valid}} = \{(i,j) \in \mathbf{M}_c \mid \mathbf{C}_c[i,j] \leq \tau_c \}
    \label{eq:DTO3}
    \end{equation}
    
    \item Combine valid matches from all classes:

    \begin{equation}
         \mathbf{M} = \bigcup_{c=1}^{N_{\mathrm{cls}}} \mathbf{M}_c^{\mathrm{valid}}
    \label{eq:DTO4}
    \end{equation}
    
    \item Define the final matched and unmatched sets:

    \begin{equation}
       \begin{aligned}
    \mathbf{m}_{\mathrm{det}} &= \{ i \mid \exists j, (i,j) \in \mathbf{M} \} \\
    \mathbf{m}_{\mathrm{tra}} &= \{ j \mid \exists i, (i,j) \in \mathbf{M} \} \\
    \mathbf{u}_{\mathrm{det}} &= \{0, \ldots, N_{\mathrm{det}}-1\} \setminus \mathbf{m}_{\mathrm{det}} \\
    \mathbf{u}_{\mathrm{tra}} &= \{0, \ldots, N_{\mathrm{tra}}-1\} \setminus \mathbf{m}_{\mathrm{tra}}
    \end{aligned}
    \label{eq:DTO5}
    \end{equation}

    \begin{itemize}
        \item \(\mathbf{m}_{\mathrm{det}} = \{ i \mid \exists j, (i,j) \in \mathbf{M} \}\)  
        Represents the set of all matched detection indices, i.e., there exists a track \(j\) matched with detection \(i\)
      
        \item \(\mathbf{m}_{\mathrm{tra}} = \{ j \mid \exists i, (i,j) \in \mathbf{M} \}\)  
        Represents the set of all matched track indices, i.e., there exists a detection \(i\) matched with track \(j\)
        
        \item \(\mathbf{u}_{\mathrm{det}} = \{0, \ldots, N_{\mathrm{det}}-1\} \setminus \mathbf{m}_{\mathrm{det}}\)  
        Represents the set of all unmatched detection indices, which is the full detection set minus the matched detection set.
        
        \item \(\mathbf{u}_{\mathrm{tra}} = \{0, \ldots, N_{\mathrm{tra}}-1\} \setminus \mathbf{m}_{\mathrm{tra}}\)  
        Represents the set of all unmatched track indices, which is the full track set minus the matched track set.
    \end{itemize}

\end{enumerate}

DTO enhances the classical Hungarian algorithm by incorporating class-aware matching and parallel processing. It independently computes the optimal assignment for each class-specific cost matrix slice, thereby ensuring precise associations at the class level. The matches are subsequently filtered using class-specific thresholds to effectively eliminate erroneous high-cost associations, which improves robustness in complex, cluttered, and occluded environments. The independent matching for each class also facilitates parallel computation, enabling real-time deployment. By aggregating valid matches across all classes, DTO generates comprehensive sets of matched and unmatched detections and tracks, where the unmatched tracks provide information for subsequent trajectory initialization and termination. This method achieves more accurate, efficient, and robust trajectory association in the challenging multi-class, multi-object scenarios commonly encountered in autonomous driving applications.

\subsection{Dynamic Motion Modeling} 

\subsubsection{Confidence Weighted Kalman Filter}

In our motion model, we introduce the confidence score from the object detector as a weighting factor in the update steps of both the \textbf{Linear Kalman Filter} and \textbf{Extended Kalman Filter}~\cite{khodarahmi2023review}. This approach assigns higher weights to detections with greater confidence during state and covariance matrix updates. Consequently, this enhancement improves the accuracy (MOTA and AMOTA) and robustness of 3D object tracking. The modified update equation for the state estimate can be expressed as

\begin{equation}
\hat{x}\_k = \hat{x}\_{k|k-1} + w_k K_k(z_k - H_k\hat{x}\_{k|k-1})
\label{eq:XK}
\end{equation}

Where $w_k$ is the confidence score of the detection at the time step $k$ and $K_k$ is the Kalman gain.

Similarly, the covariance update is adjusted to:

\begin{equation}
P_k = (I - w_kK_kH_k)P_{k|k-1}(I - w_kK_kH_k)^T + w_k^2K_kR_kK_k^T
\label{eq:PK}
\end{equation}

This weighted approach effectively incorporates the reliability of detections into the filtering process, leading to more accurate and robust 3D object tracking performance.

\begin{table*}
    \begin{center}
    \caption{ 
    A comparison between our method and other methods on the nuScenes test set.   
    $\ddagger$ means the GPU device.
    }
        \label{table:nu_test}
        \setlength{\tabcolsep}{1.6mm}
        {
        \begin{tabular}{cccc|ccc|ccc}
        \toprule
        \multicolumn{1}{c}{\textbf{Method}} & \textbf{Device} & \textbf{Detector} & \textbf{Input} & \textbf{AMOTA}$\uparrow$ & \textbf{MOTA}$\uparrow$ & \textbf{FPS}$\uparrow$ & \textbf{IDS}$\downarrow$ & \textbf{FN}$\downarrow$ & \textbf{FP}$\downarrow$ \\ \midrule
                                     EagerMOT~\cite{kim2021eagermot}               & \textbf{\text{--}}       & CenterPoint~\cite{yin2021center}\&Cascade R-CNN~\cite{cai2018cascade}         & 2D+3D       & 67.7      & 56.8     & 4    & 1156    & 24925   & 17705   \\
                                   CBMOT~\cite{benbarka2021score}     & I7-9700          & CenterPoint~\cite{yin2021center}\&CenterTrack~\cite{zhou2020tracking}          & 2D+3D         & 67.6      & 53.9      & \textcolor{red}{80.5}    & 709    & 22828   & 21604   \\
                                   Minkowski~\cite{gwak2022minkowski} & TITAN$\ddagger$  & Minkowski~\cite{gwak2022minkowski}         & 3D      & 69.8      & 57.8     & 3.5    & 325    & 21200   & 19340   \\
                                   ByteTrackv2~\cite{zhang2023bytetrackv2} & \textbf{\text{--}}       & TransFusion-L~\cite{bai2022transfusion}         & 3D      & 70.1      & 58     & \textbf{\text{--}}    & 488    & 21836   & 18682   \\
                                   3DMOTFormer~\cite{ding20233dmotformer}& 2080Ti$\ddagger$       & BEVFuison~\cite{liu2023bevfusion}       & 2D+3D      & 72.5      & 60.9     & \textcolor{blue}{54.7}    & 593    & 20996   & \textcolor{blue}{17530}   \\  
                                   Poly-MOT~\cite{li2023poly}               & 9940X       & LargeKernel3D~\cite{chen2022scaling}         &2D+3D       & \textcolor{blue}{75.4}      & \textcolor{blue}{62.1}     & 3    & \textcolor{blue}{292}    & \textcolor{blue}{17956}   & 19673   \\ 
                                  Fast-Poly~\cite{li2024fast}               & 7945HX       & LargeKernel3D~\cite{chen2022scaling}         &2D+3D       & \textcolor{blue}{75.8}      & \textcolor{blue}{62.8}     & 34.2    & \textcolor{blue}{326}    & \textcolor{blue}{18415}   &  \textcolor{blue}{17098}  \\ \midrule

                                  \textbf{Easy-Poly (Ours)}               & 4090$\ddagger$       & LargeKernel3D~\cite{chen2022scaling}         &2D+3D       &  \textcolor{red}{76.3}      & \textcolor{red}{63.5}     & \textcolor{blue}{34.8}    & \textcolor{red}{270}    & \textcolor{red}{16602}   &   \textcolor{red}{16319}  \\
        \bottomrule
        \end{tabular}}
        \end{center}
 \end{table*}

\subsubsection{Adaptive Noise Covariances}

We have introduced novel dynamic adjustments to the noise covariance matrices, specifically the measurement noise covariance $R$ and the process noise covariance $Q$. These adjustments are implemented in both the \textbf{Linear Kalman Filter} and the \textbf{Extended Kalman Filter}. The dynamic parameter tuning enhances the adaptability and robustness of the system while optimizing the balance between accuracy and response speed. Experimental results indicate that an adjustment range of 10\% for $R$ and $Q$, corresponding to scaling factors of 0.9 and 1.1, yields optimal performance. This calibrated strategy preserves system stability through moderate parameter modulation. In contrast, more aggressive adjustments, such as scaling factors of 0.8 or 1.2, may cause system instability or excessive response, despite potentially enabling faster adaptation to significant changes.

The dynamic adjustment of the measurement noise covariance is based on the magnitude of the measurement residual, which is used to adaptively modify the value of $R$. Specifically, when the measurement residual is small, the value of $R$ is decreased; conversely, when the measurement residual is large, the value of $R$ is increased. The specific formula is as follows:

\begin{equation}
R_{new} = \begin{cases}
0.9 \cdot R, & \text{if } \|res\| < 1.0 \\
1.1 \cdot R, & \text{if } \|res\| > 5.0 \\
R, & \text{otherwise}
\end{cases}
\label{eq:R}
\end{equation}

\(R_{new}\) is the adjusted measurement noise covariance, \(R\) is the original measurement noise covariance, \(\|res\|\) represents the Euclidean norm (L2 norm) of the measurement residual, \(\|res\| = \sqrt{\sum_{i=1}^n res_i^2}\), where \(res_i\) is the i-th component of the residual vector. This formula expresses the following logic:
If the residual norm is less than 1.0, the value of $R$ is decreased by 10\%. If the residual norm exceeds 5.0, the value of $R$ is increased by 10\%. When the residual norm lies between 1.0 and 5.0, the value of $R$ remains unchanged. This dynamic adjustment strategy is designed to modulate the behavior of the Kalman filter according to the reliability of the measurements. Specifically, a small measurement residual leads to increased confidence in the measurement by reducing $R$, whereas a large measurement residual results in decreased confidence by increasing $R$.

The dynamic adjustment of the process noise covariance is based on the current state, allowing adaptive modification of the process noise covariance $Q$. For instance, the value of $Q$ is adjusted according to the magnitude of velocity. The formula is as follows:

\begin{equation}
Q_{new} = \begin{cases}
0.9 \cdot Q, & \text{if } \|\mathbf{v}\| < 1.0 \\
1.1 \cdot Q, & \text{if } \|\mathbf{v}\| > 10.0 \\
Q, & \text{otherwise}
\end{cases}
\label{eq:Q}
\end{equation}

The \(Q_{new}\) is the noise covariance, \(Q\) is the original process noise covariance, \(\mathbf{v} = [state_1, state_2]\) represents the first two components of the state vector (assumed to be velocity components), \(\|\mathbf{v}\| = \sqrt{state_1^2 + state_2^2}\) is the Euclidean norm (L2 norm) of the velocity vector.
This formula expresses the following logic: If the velocity norm is less than 1.0, Q is reduced by 10\%. If the velocity norm is greater than 10.0, Q increases by 10\%. If the velocity norm is between 1.0 and 10.0, Q remains unchanged.

This dynamic adjustment strategy is designed to modify the process model of the Kalman filter based on the current state, specifically the velocity. When the velocity is low, the process noise covariance $Q$ is decreased, reflecting increased confidence in the system dynamics; conversely, when the velocity is high, $Q$ is increased, indicating reduced confidence in the system dynamics.

In complex autonomous driving scenarios with dense traffic, frequent occlusions, and small or fast-moving objects, velocity-dependent adjustment of the process noise covariance aligns noise characteristics with target dynamics, improving tracking accuracy. Adaptive tuning of the measurement noise covariance addresses sensor noise variability from vibration and partial visibility, ensuring robust sensor fusion. Range-dependent modulation of both measurement and process noise covariances further refines precision for nearby targets while accommodating greater uncertainty for distant ones. These adaptive strategies collectively enhance filter robustness in tracking multiple targets amid cluttered, occluded, and dynamic environments.
 
\begin{table*}
\vspace{0.5em}
\begin{center}
\caption{
{A comparison of existing methods applied to the nuScenes val set.}}
\label{table:nu_val}
\setlength{\tabcolsep}{2.4mm}
{
\begin{tabular}{cccccccc}
\toprule
\bf{Method} & \bf{Detector} & \bf{Input Data} & \bf{MOTA$\uparrow$} & \bf{AMOTA$\uparrow$} & \bf{AMOTP$\downarrow$} & \bf{FPS$\uparrow$} & \bf{IDS$\downarrow$}  \\ \hline
CBMOT~\cite{benbarka2021score}   & CenterPoint~\cite{yin2021center} \& CenterTrack~\cite{zhou2020tracking} & 2D + 3D & - & 72.0 & \textbf{\textcolor{red}{48.7}}  & \textbf{\textcolor{red}{80}} & 479   \\
EagerMOT~\cite{kim2021eagermot}  & CenterPoint~\cite{yin2021center} \& Cascade R-CNN~\cite{cai2018cascade} & 2D + 3D  & -- & 71.2   & 56.9 & 13  & 899    \\
SimpleTrack~\cite{pang2022simpletrack}  & CenterPoint~\cite{yin2021center} & 3D & 60.2 & 69.6  & 54.7 & 0.5  & 405  \\
CenterPoint~\cite{yin2021center}   & CenterPoint~\cite{yin2021center} & 3D & -- & 66.5  & 56.7 & -- & 562 \\ 
OGR3MOT~\cite{zaech2022learnable}  & CenterPoint~\cite{yin2021center} &3D & 60.2 & 69.3  & 62.7 & 12.3 & \textbf{\textcolor{blue}{262}}  \\ \hline

\textbf{Poly-MOT}~\cite{li2023poly}      & CenterPoint~\cite{yin2021center} & 3D & 61.9 & 73.1   & \textbf{\textcolor{blue}{52.1}} & 5.6 & 281   \\ 

\textbf{Poly-MOT}~\cite{li2023poly}      & LargeKernel3D-L~\cite{chen2022scaling}  & 3D & 54.1 & \textbf{\textcolor{blue}{75.2}}    & 54.1  & 8.6 & 354 \\

\textbf{Fast-Poly}~\cite{li2024fast}      & CenterPoint~\cite{yin2021center} & 3D & \textbf{\textcolor{blue}{63.2}}  & 73.7   &  -- & 28.9 & 414   \\ \hline
 
\textbf{Easy-Poly (Ours)}       & CenterPoint~\cite{yin2021center} & 2D + 3D & \textbf{\textcolor{blue}{64.4}} & \textbf{\textcolor{blue}{74.5}}   & 54.9  & \textbf{\textcolor{blue}{34.6}} & \textbf{\textcolor{blue}{272}}   \\
\textbf{Easy-Poly (Ours)}       & LargeKernel3D~\cite{chen2022scaling}  & 2D + 3D  & \textbf{\textcolor{red}{65.5}} & \textbf{\textcolor{red}{75.6}}    & \textbf{\textcolor{blue}{53.6}}  & \textbf{\textcolor{blue}{34.9}}  & \textbf{\textcolor{red}{242}} \\ \hline
\end{tabular}}
\end{center}
\end{table*}

\subsection{Life-cycle Adjustment}

Threshold adjustment plays a critical role in the performance of object detection and tracking. The \textbf{Intersection over Union (IoU)} threshold governs the balance between precision and recall. A higher threshold promotes more accurate detections but may fail to identify objects with slightly lower IoU values. In contrast, a lower threshold increases sensitivity, potentially detecting a greater number of objects at the expense of a higher false positive rate. Through extensive experimentation, this threshold is iteratively optimized to achieve an optimal trade-off between detection accuracy and false alarm rate, thereby improving overall system performance in autonomous driving contexts.

During the life-cycle management stage, Easy-Poly demonstrates improvements over the baseline Fast-Poly with respect to the motion model. Specifically, the wheelbase ratio and rear tire ratio parameters were adjusted, with optimal values identified via grid search or Bayesian optimization, changing from the original 0.8 and 0.5 to 0.6 and 0.3, respectively. These modifications enhance tracking performance and robustness. Furthermore, the maximum age parameter was increased to 20 across all detection and tracking categories. This adjustment substantially prolongs the duration of object tracking while preserving high tracking accuracy and computational efficiency. As a result, a significant reduction in frame loss during life-cycle management was observed, yielding more complete and robust tracking trajectories for each object of interest.

\section{Experiments}
\label{sec: experiments}

A series of experiments were conducted to validate the effectiveness of Easy-Poly.

\subsection{Implementation Details}

The tracking framework is implemented in Python and runs on an NVIDIA RTX 4090 GPU. The hyperparameters were optimized to maximize AMOTA in the nuScenes dataset~\cite{caesar2020nuscenes} test set. The Category-specific and SF thresholds for nuScenes are set as follows: \textit{Bic} 0.14, \textit{Car} 0.16, \textit{Moto} 0.16, \textit{Bus} 0.12, \textit{Tra} 0.13, \textit{Tru} 0, and \textit{Ped} 0.13. The NMS threshold is uniformly fixed at 0.08 across all categories and datasets. Scale-NMS~\cite{huang2021bevdet} is applied to the \textit{Bic} and \textit{Ped} categories on nuScenes. In addition to the default NMS, a novel similarity metric is introduced for the \textit{Bic}, \textit{Ped}, \textit{Bus}, and \textit{Tru} categories. The confidence-based decay rates for nuScenes are: \textit{Ped} 0.18, \textit{Car} 0.26, \textit{Tru} and \textit{Moto} 0.28, \textit{Tra} 0.22, and \textit{Bic} and \textit{Bus} 0.24.

\begin{table}
        \caption{A comparison of different data association algorithms is conducted using the CenterPoint method (lines 1--7) and the LargeKernel3D method (lines 8--11) on the nuScenes validation set. Among these, lines 1--3 correspond to the Fast-Poly framework, while lines 4--11 represent the latest version of the Easy-Poly framework.}

        \label{tab:nus_assoc}
        \setlength{\tabcolsep}{0.9mm}
        \begin{tabular}{cccccc}
        \toprule
        \multicolumn{1}{c}{\textbf{Algorithms}} & \textbf{MOTA}$\uparrow$ & \textbf{AMOTA}$\uparrow$ & \textbf{AMOTP}$\downarrow$ & \textbf{IDS}$\downarrow$ & \textbf{FN}$\downarrow$\\ 
        
        \midrule
         MNN  & 62.2  & 72.5 & 52.4  & 433 & 16644  \\  
         Greedy    & 62.3  & 72.7 &  53.4  & 428  & 17647 \\
         Hungarian & \textbf{\textcolor{blue}{63.2}}  & \textbf{\textcolor{blue}{73.7}} & \textbf{\textcolor{blue}{52.1}} & \textbf{\textcolor{blue}{414}}  & \textbf{\textcolor{blue}{15996}}  \\
        
        \midrule  
        \textbf{MNN}  & 63.7  & 73.6 & 54.8  & 406 &  \textbf{\textcolor{red}{15873 }}\\  
        \textbf{Greedy}    &  64.0  & 73.7  &  54.6  & 368  & 16736 \\
        \textbf{Hungarian}  & 64.3  & 74.3 & \textbf{\textcolor{red}{54.3}} &  335 & 16892  \\

        \textbf{DTO}   & \textbf{\textcolor{red}{64.4}}  &   \textbf{\textcolor{red}{74.5}} & 54.9 &  \textbf{\textcolor{red}{272}} & 16982  \\

         \midrule
         \textbf{MNN}  & 64.1  & 73.9 & 54.0  & 370 & 15865  \\  
         \textbf{Greedy}    & 64.5  & 74.3 & 53.7  & 307  & 16014 \\
        \textbf{Hungarian}  & 64.7  & 74.8 & 53.9  & 291  & 15923 \\

        \textbf{DTO}  & \textbf{\textcolor{red}{65.5}}  & \textbf{\textcolor{red}{75.6}} & \textbf{\textcolor{red}{53.6}}  & \textbf{\textcolor{red}{242}}  & \textbf{\textcolor{red}{15481}} \\

    \bottomrule
\end{tabular}
\end{table}

\subsection{Comparative Evaluations} 
\label{sec: Comparative}

Easy-Poly achieves a \textbf{76.3\%} AMOTA on the test set, outperforming most existing 3D MOT methods. As shown in Table \ref{table:nu_test}, it attains a low IDS count of \textbf{270} while maintaining the highest AMOTA (\textbf{76.3\%}) among all modal methods, demonstrating stable tracking without recall loss. With minimal computational overhead, Easy-Poly delivers strong results suitable for real-world autonomous driving applications. Its False Negative and False Positive rates further confirm robust continuous tracking and high recall. The reduced IDS and FP values highlight its effectiveness in tracking small objects and maintaining performance under complex scenarios and adverse weather. These results collectively demonstrate the algorithm’s robustness across diverse challenging environments.

On the nuScenes validation set (Table \ref{table:nu_val}), Easy-Poly attains \textbf{75.6\%} AMOTA at \textbf{34.9} FPS, surpassing existing methods. Using the same detector, it outperforms Fast-Poly \cite{li2024fast} on nearly all key metrics, with notable improvements in accuracy (\textbf{+1.8\%} AMOTA, \textbf{+2.3\%} MOTA) and speed (\textbf{+6.0} FPS). Although slightly slower than CBMOT \cite{benbarka2021score} and 3DMOTFormer \cite{ding20233dmotformer}, Easy-Poly achieves substantially higher accuracy while preserving real-time capability. Its open-source implementation provides a robust baseline for future 3D MOT research.

Table~\ref{tab:nus_assoc} evaluates association algorithms, showing Easy-Poly outperforms Fast-Poly and LargeKernel3D exceeds CenterPoint across all metrics. DTO achieves the highest performance within LargeKernel3D, improving AMOTA by 0.8 points (75.6\% vs. 74.8\%), reducing AMOTP to 53.6, IDS by 49 (242 vs. 291), and FN by 442 (15,481 vs. 15,923) compared to the Hungarian algorithm. These gains demonstrate DTO’s robustness against occlusions, false positives, and complex dynamics, effectively preserving tracking continuity and accuracy in dense, small-object scenarios. Moderate AMOTP improvement suggests potential for enhanced localization via advanced spatiotemporal fusion.

Table~\ref{tab:nu_confidence} shows Easy-Poly attains \textbf{75.6\%} AMOTA, a \textbf{1.3\%} gain over Count \& Max-age, with MOTA at \textbf{65.5\%} and FN at \textbf{15,481}, the lowest among methods, while maintaining \textbf{34.9 FPS}. Integrating confidence scores with count-based life-cycle management optimizes object existence and tracklet termination, reducing missed detections. The model exhibits robustness to occlusions and appearance changes, enhancing small object detection through fine-grained confidence evaluation and multi-frame fusion, thereby stabilizing continuous tracking.

\subsection{Ablation Studies}

\begin{table}
        \caption{Comparison of life-cycle modules on the nuScenes validation set.  
        \textbf{Average} indicates deletion based on the online average score.  
        \textbf{Latest} indicates deletion based on the most recent score.  
        \textbf{Max-age} indicates deletion based on the duration of continuous mismatches.  
        All other settings are under the best performance.
        }
        \label{tab:nu_confidence}
        \setlength{\tabcolsep}{0.6mm}
        \begin{tabular}{ccccc}
        \toprule
       \multicolumn{1}{c}{\textbf{Strategy}} & \textbf{AMOTA}$\uparrow$ & \textbf{MOTA}$\uparrow$ & \textbf{FPS}$\uparrow$ & \textbf{FN}$\downarrow$ \\  
        
        \midrule
        \textbf{Count \& Max-age}          & 74.3      & 63.8     & 34.3  & 16024    \\
         \textbf{Confidence \& Latest}        & 71.8      & 63.7     & \textbf{\textcolor{red}{50.2}} & 17907   \\
         \textbf{Confidence \& Average}       & 74.6      & 64.2     & 35.6 & 16063   \\
         \textbf{Confidence(Ours) \& Average }      & \textbf{\textcolor{red}{75.6}}     & \textbf{\textcolor{red}{65.5}}     & 34.9  & \textbf{\textcolor{red}{15481}}  \\
        
    \bottomrule
\end{tabular}
\end{table}

\begin{table}
        \caption{The results of the ablation study of each module on the NuScenes val set. \textbf{OS} means original state. \textbf{CNMSMM} means 3D Detection Module. \textbf{DTO} means Dynamic Track-Oriented Data Association Module. \textbf{DMM} means Dynamic Motion Modeling  Module. \textbf{LM} means Life-cycle Management Module.}
        \label{tab:ablation_study_module}
        \setlength{\tabcolsep}{0.01mm}
        \begin{tabular}{ccccc}
        \toprule
       \multicolumn{1}{c}{\textbf{Module}} & \textbf{AMOTA}$\uparrow$ & \textbf{IDS}$\uparrow$ & \textbf{FN}$\uparrow$ & \textbf{FP}$\downarrow$ \\  
        \midrule
        \textbf{Os}                  & 73.2   & 413    & 18042 & 18009\\
        \textbf{Os + CNMSMM}             & 73.8   & 324    & 17293 & 17498\\
        \textbf{Os + CNMSMM + DTO}        & 74.2   & 276   & 16816 & 17140\\
        \textbf{Os + CNMSMM + DTO + DMM}       & 74.9   & 288    & 16171 & 16532\\
        \textbf{Os + CNMSMM + DTO + DMM + LM}  & \textbf{\textcolor{red}{75.6}}  & \textbf{\textcolor{red}{242}}   & \textbf{\textcolor{red}{15481}} & \textbf{\textcolor{red}{13270}} \\
    \bottomrule
\end{tabular}
\end{table}

\begin{figure*}[t]
    \centering
    \includegraphics[width=0.98\linewidth]{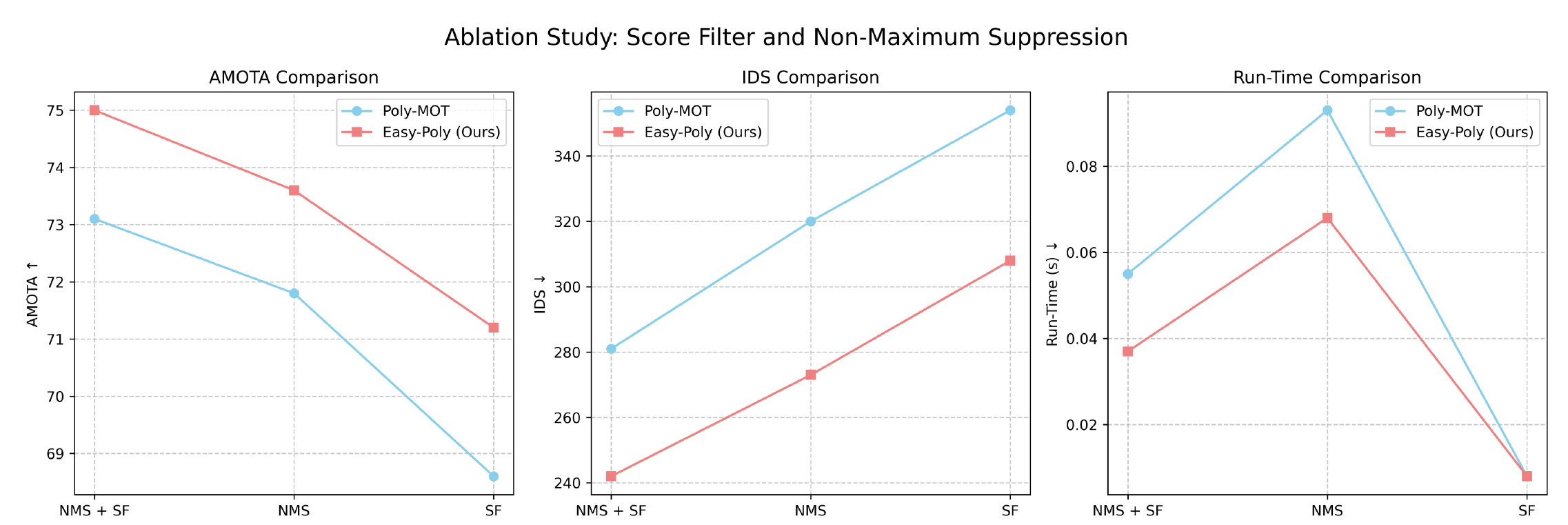}
    \vspace{-8pt}
    \caption
    {
      This ablation study evaluates the effects of incorporating the Score Filter and Non-Maximum Suppression on performance, with run-time measured as the execution duration of the pre-processing module. The analysis compares the baseline Poly-MOT method with the proposed Easy-Poly approach.
     }
     \Description{}
    \label{fig: NMSFS}
\end{figure*}

\begin{figure*}[t]
    \centering
    \includegraphics[width=0.98\linewidth]{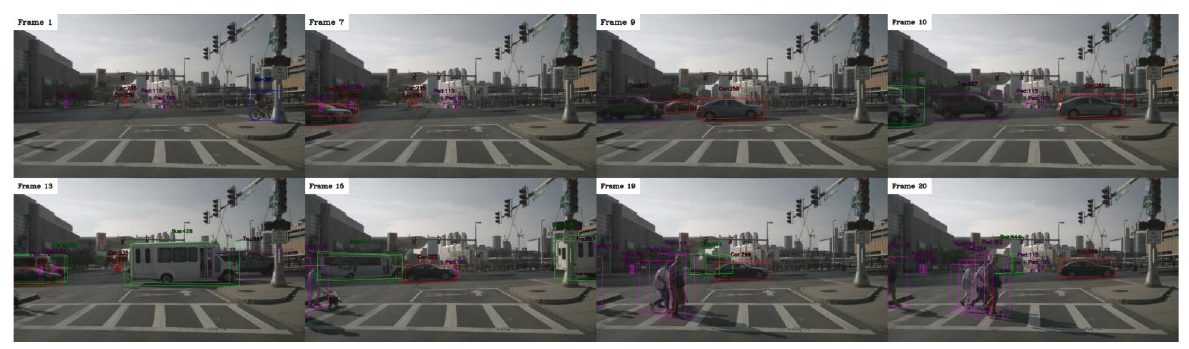}
    \vspace{-8pt}
    \caption
    {
      Visualization of the Easy-Poly framework for 3D detection and multi-object tracking performance at a traffic intersection is presented. A 20-frame tracking sequence is used to evaluate the robustness of detection and tracking under challenging conditions, including small objects, distant targets, and crowded scenes. The tracked object categories include bicycles, buses, cars, pedestrians, and trucks.
     }
     \Description{}
    \label{fig: Visualization}
\end{figure*}

The ablation experiment presented in Fig~\ref{fig: NMSFS} validates the integration of the Score Filter (SF) with Non-Maximum Suppression (NMS) within Easy-Poly, achieving an AMOTA of \textbf{75\%} under the combined NMS + SF setting. This performance surpasses that of NMS-only and SF-only configurations by \textbf{1.4\%} and \textbf{3.8\%}, respectively, indicating enhanced detection precision and tracklet reliability. Easy-Poly reduces identity switches (IDS) to \textbf{242} with NMS + SF, outperforming Poly-MOT’s \textbf{281} IDS, which reflects improved robustness against occlusion and appearance variations. Runtime efficiency is also improved, with Easy-Poly processing frames in \textbf{0.037 s} compared to Poly-MOT’s \textbf{0.055 s}, achieving a favorable balance between speed and accuracy suitable for real-time deployment. These results confirm the superiority of Easy-Poly in detection accuracy, identity consistency, and computational efficiency, advancing multi-object tracking in autonomous driving through the synergistic application of SF and NMS.

Table~\ref{tab:ablation_study_module} demonstrates the incremental improvements resulting from the integration of CNMSMM, DTO, DMM, and LM modules within Easy-Poly on the NuScenes dataset. The complete model increases AMOTA from 73.2\% to \textbf{75.6\%} (+2.4\%) and reduces IDS from 413 to \textbf{242}, evidencing enhanced identity preservation and tracking stability. False negatives and false positives decrease by \textbf{2,561} and \textbf{4,739}, respectively, underscoring improved detection precision and robustness, particularly for small and occluded objects. This modular synergy enhances detection accuracy, association reliability, and temporal consistency, confirming Easy-Poly’s advancement in precision and robustness for multi-object tracking in autonomous driving.

\subsection{Visualization}

Fig~\ref{fig: Visualization} illustrates qualitative tracking results of Easy-Poly in a complex urban intersection, highlighting its robust 3D multi-modal detection and tracking capabilities under safety-critical scenarios. The framework consistently achieves precise detection and continuous tracking of both near-field and distant small objects; pedestrians with IDs 115, 166, and 185 are tracked uninterruptedly from Frame 1 to Frame 20 without missed detections or identity switches, while the distant vehicle Car:259 is reliably tracked from Frame 1 to Frame 13, demonstrating effective long-term association for far-field targets. Easy-Poly further demonstrates robustness in crowded environments by accurately detecting and preserving identities amid pedestrian congestion and partial occlusions at Frame 20, addressing a key challenge in multi-object tracking. Additionally, it maintains stable tracking performance under occlusions and complex vehicle interactions, as exemplified by bus ID 428, which is tracked from Frame 10 to Frame 15 despite partial occlusions. These results confirm the superiority of Easy-Poly over prior approaches, such as Fast-Poly, in continuous and accurate multi-object tracking of small, distant, and densely clustered targets in real-world urban driving scenarios.

\section{Conclusion}
\label{sec:conclusion}

This paper introduces Easy-Poly, a filter-based 3D multi-object tracking framework that substantially enhances detection and tracking accuracy and robustness through the integration of innovative Camera-LiDAR fusion detection (CNMSMM), DTO, and DMM strategies. Easy-Poly achieves superior performance, particularly in small object detection, occlusion handling, and complex dynamic environments. Experimental results demonstrate that it outperforms current state-of-the-art methods on both mAP and AMOTA metrics while maintaining real-time operation. Although Easy-Poly makes significant advances in multi-modal fusion and motion modeling, opportunities remain to improve sustained tracking under extreme weather conditions and prolonged occlusions. Future work will concentrate on enhancing the model's environmental adaptability and the deep integration of temporal information from multiple sensors to advance 3D multi-object tracking toward greater intelligence and stability, thereby facilitating safer and more reliable perception and decision-making in autonomous driving systems.

\balance
\bibliographystyle{plain}
\bibliography{main}


\end{document}